\documentclass[acmtog]{acmart}
\AtBeginDocument{%
  }

\usepackage{multirow}
\usepackage{enumitem}

\copyrightyear{2025}
\acmYear{2025}
\setcopyright{acmlicensed}\acmConference[SA Conference Papers '25]{SIGGRAPH Asia 2025 Conference Papers}{December 15--18, 2025}{Hong Kong, Hong Kong}
\acmBooktitle{SIGGRAPH Asia 2025 Conference Papers (SA Conference Papers '25), December 15--18, 2025, Hong Kong, Hong Kong}
\acmDOI{10.1145/3757377.3763951}
\acmISBN{979-8-4007-2137-3/2025/12}
\acmSubmissionID{1949}

\begin{document}

\title{PhysHMR: Learning Humanoid Control Policies from Vision for Physically Plausible Human Motion Reconstruction}

\author{Qiao Feng}
\email{fengqiao@seas.upenn.edu}
\orcid{0000-0003-0625-651X}
\affiliation{%
  \institution{University of Pennsylvania}
  \city{Philadelphia}
  \state{Pennsylvania}
  \country{USA}
}

\author{Yiming Huang}
\email{ymhuang9@seas.upenn.edu}
\orcid{0009-0004-4001-0630}
\affiliation{%
  \institution{University of Pennsylvania}
  \city{Philadelphia}
  \state{Pennsylvania}
  \country{USA}
}

\author{Yufu Wang}
\email{yufu@seas.upenn.edu}
\orcid{0000-0001-9907-8382}
\affiliation{%
  \institution{University of Pennsylvania}
  \city{Philadelphia}
  \state{Pennsylvania}
  \country{USA}
}

\author{Jiatao Gu}
\email{jgu32@cis.upenn.edu}
\orcid{0000-0003-3578-2711}
\affiliation{%
  \institution{University of Pennsylvania}
  \city{Philadelphia}
  \state{Pennsylvania}
  \country{USA}
}

\author{Lingjie Liu}
\email{lingjie.liu@seas.upenn.edu}
\orcid{0000-0003-4301-1474}
\affiliation{%
  \institution{University of Pennsylvania}
  \city{Philadelphia}
  \state{Pennsylvania}
  \country{USA}
}

\renewcommand{\shortauthors}{Qiao et al.}

\begin{abstract}
Reconstructing physically plausible human motion from monocular videos remains a challenging problem in computer vision and graphics. Existing methods primarily focus on kinematics-based pose estimation, often leading to unrealistic results due to the lack of physical constraints. To address such artifacts, prior methods have typically relied on physics-based post-processing following the initial kinematics-based motion estimation. However, this two-stage design introduces error accumulation, ultimately limiting the overall reconstruction quality.
In this paper, we present PhysHMR, a unified framework that directly learns a visual-to-action policy for humanoid control in a physics-based simulator, enabling motion reconstruction that is both physically grounded and visually aligned with the input video. A key component of our approach is the pixel-as-ray strategy, which lifts 2D keypoints into 3D spatial rays and transforms them into global space.
These rays are incorporated as policy inputs, providing robust global pose guidance without depending on noisy 3D root predictions. This soft global grounding, combined with local visual features from a pretrained encoder, allows the policy to reason over both detailed pose and global positioning. To overcome the sample inefficiency of reinforcement learning, we further introduce a distillation scheme that transfers motion knowledge from a mocap-trained expert to the vision-conditioned policy, which is then refined using physically motivated reinforcement learning rewards. Extensive experiments demonstrate that PhysHMR produces high-fidelity, physically plausible motion across diverse scenarios, outperforming prior approaches in both visual accuracy and physical realism.
\end{abstract}


\begin{teaserfigure}
\centering
\includegraphics[width=\textwidth]{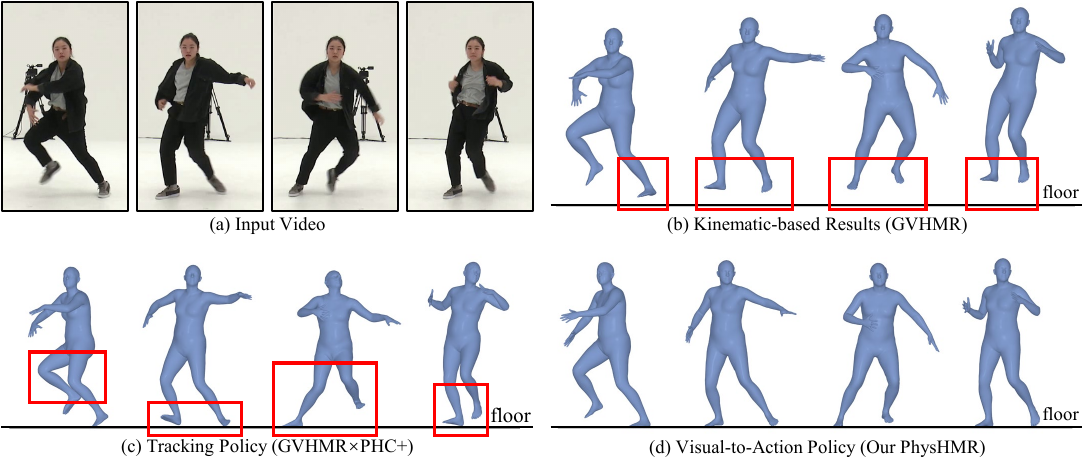}
\caption{
Given a monocular video (a), (b) kinematic-based methods (e.g., GVHMR~\cite{shen2024gvhmr}) often cannot produce physically plausible results and suffer from artifacts like foot floating. (c) While tracking-based controllers (e.g., PHC+~\cite{Luo2023PerpetualHC}) can enforce physical plausibility, they may amplify errors from inaccurate motion reconstruction, leading to unnatural behaviors. (d) In contrast, our PhysHMR model learns a visual-to-action policy that directly predicts control signals from visual input, preventing error amplification and producing motions that are both physically plausible and visually aligned with the input video (a). As videos are the most effective way to assess the physical plausibility of the results, we encourage readers to view our supplementary video. 
}
  \Description{teaser}
  \label{fig:teaser}
\end{teaserfigure}


\maketitle
\section{Introduction}
Faithfully reconstructing human body dynamics from monocular videos, also known as Human Mesh Recovery (HMR), is a fundamental problem in computer vision, graphics, and robotics. Recent advances in human motion reconstruction~\cite{TRAM, WHAM, ye2023slahmr, shen2024gvhmr, PHALP, yuan2022glamr, TRACE} have achieved high accuracy in estimating body pose and shape. However, most existing methods overlook physically plausible body dynamics, leading to various artifacts such as foot sliding, ground penetration, and inconsistent contact behavior (see Fig.~\ref{fig:teaser}(b)). Achieving physically plausible human motion reconstruction remains an open and challenging problem.

Prior works have attempted to introduce physical constraints through a post-hoc correction stage. Some approaches incorporate analytical priors derived from rigid body dynamics, such as the Euler-Lagrange equation~\cite{zhang2024physpt,jiang2023drop,zhang2024incorporating}, while others leverage reinforcement learning to train humanoid controllers that track pre-reconstructed motion~\cite{yuan2021simpoe}. Although these methods improve physical realism to some extent, they share a \emph{common limitation}: motion is first reconstructed from visual cues alone and then refined by a separate physics module. This decoupled design overlooks the ambiguity inherent in monocular videos, where multiple plausible motions can explain the same visual observation. Once a single solution is selected in the reconstruction stage, the downstream physics module can no longer access the full observational context, leading to suboptimal corrections and limited consistency with the visual evidence (see Fig.~\ref{fig:teaser}c).

In light of these limitations, we argue that a more effective approach is to {\bf unify motion estimation and physical reasoning within a single framework}, allowing visual cues and physical constraints to inform the same decision process. To this end, we propose PhysHMR, a novel framework that directly learns a visual-to-action policy to control a simulated humanoid directly from monocular video observations, resulting in reconstructed motion that is both visually consistent and physically plausible. Unlike prior two-stage approaches, PhysHMR unifies the two stages through a single policy network that jointly reasons over visual observations and physical dynamics. By executing motion within a physics-based simulator~\cite{makoviychuk2021isaac}, it naturally enforces physical constraints such as ground contact, joint limits, and momentum conservation. By conditioning the policy directly on image features, we can exploit rich visual context beyond skeletal pose estimations, enabling the humanoid to produce motion that faithfully aligns with the input video while adhering to physical laws.

Training high-dimensional visual-control policies purely with reinforcement learning is often sample-inefficient and unstable \cite{Luo_2024_CVPR}. To address these issues, PhysHMR proposes a distillation strategy that transfers knowledge from a mocap-trained imitation expert, thereby facilitating the training of the visual-to-action policy. Specifically, a pretrained visual encoder\cite{shen2024gvhmr} extracts features from each video frame, which serve as local pose references for the control policy. These features retain rich pose information without committing to potentially inaccurate 3D reconstructions. The expert controller, trained on high-quality motion capture data, provides action supervision that imparts strong human motion priors, which significantly accelerates convergence and stabilizes learning. The policy is further refined with reinforcement learning, using a composite reward that balances motion imitation, realism through adversarial motion priors, and physical smoothness.

Since physical plausibility must be assessed in the global pose space rather than the local pose space, it is necessary to estimate global pose information (i.e., the root joint position) in addition to local pose references from images. However, predicting the 3D root joint position from monocular video is often noisy, which significantly compromises the robustness of policy generalization. This is because inconsistencies between local pose estimates and erroneous 3D root predictions can lead to unnatural global motions—for example, the local pose may indicate forward movement, while the noisy root prediction pulls the motion backward, resulting in jittery or unstable behavior. Such mismatches make it difficult for the policy to produce physically consistent dynamics in global space. To address this, instead of relying on explicit 3D root prediction, we lift multiple detected 2D keypoints into 3D rays, which serve as a soft global pose reference. These spatial rays condition the policy to predict actions that transform the humanoid into globally consistent poses without requiring strict absolute 3D root input. This approach provides gentle global information, improves the robustness of policy execution, and enables physically plausible human motion reconstruction. 

We evaluate PhysHMR on challenging motion datasets, including Human3.6M, AIST++, and EMDB2, showing comparable motion accuracy to state-of-the-art kinematics-based methods while significantly improving physical plausibility. Our approach reduces common non-physical artifacts (e.g. foot sliding, ground penetration), improving the suitability of reconstructed motion for downstream applications such as simulation, animation, and robotics. 

In summary, our contributions are three-fold:
\begin{itemize}[leftmargin=*]
\item We present PhysHMR, the {\bf first} unified framework for jointly performing human motion perception and control, enabling high-quality and physically plausible human motion reconstruction from monocular videos. 

\item We introduce a distillation approach to distill a visual-to-action policy from a pretrained mocap imitation policy, which accelerates convergence and stabilizes policy learning. 

\item We propose a soft global grounding strategy by lifting 2D keypoints into 3D spatial rays, avoiding the need for noisy 3D root predictions and enabling robust policy learning of physically plausible motion in global space.
\end{itemize}

\section{Related Works}
\begin{figure*}[t]
    \centering
\includegraphics[width=0.9\linewidth]{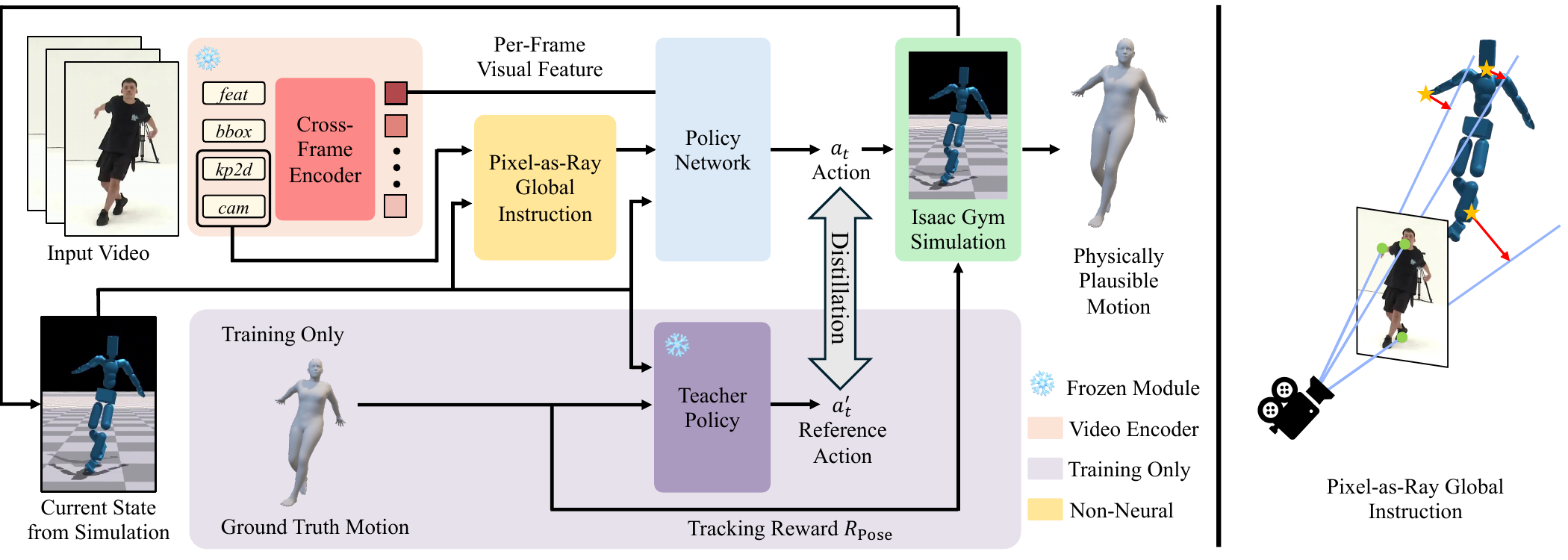}
\caption{
Overview of our pipeline. A visual-to-action policy reconstructs physically plausible motion from monocular videos. Training efficiency is improved by combining reinforcement learning and knowledge distillation. Global motion is guided using a \textit{pixel-as-ray} module that lifts 2D keypoints into 3D rays.
}
  \Description{pipeline}
    \label{fig:pipeline}
\end{figure*}
\subsection{Kinematics-Based Human Mesh Recovery}
Parametric human models~\cite{SMPL:2015, STAR:2020, SMPL-X:2019, 50649} have been widely adopted to reconstruct human motion from monocular video. 
Early works~\cite{10.1007/978-3-319-46454-1_34, Arnab_CVPR_2019, MuVS:3DV:2017, xiang2019monocular} focus on fitting these models to individual image frames. 
More recently, regression-based approaches, which leverage large-scale datasets, have gained attention for their ability to achieve general-purpose human mesh recovery~\cite{goel2023humans, cai2023smplerx, yin2025smplest}. 
To account for dynamic camera movements, data-driven methods have been extended to estimate per-frame camera poses~\cite{TRACE, WHAM, yuan2022glamr}. Additionally, SLAM (Simultaneous Localization and Mapping) techniques have proven effective for robust camera motion estimation, further enhancing human motion recovery in complex scenarios~\cite{TRAM}.
HuMoR~\cite{rempe2021humor} learns a generative motion prior that improves temporal consistency and robustness in pose estimation.
Despite these advances in human mesh recovery, purely kinematic methods often exhibit artifacts like foot sliding, ground penetration, and momentum inconsistency.

To address such artifacts, prior works have used physical priors as an auxiliary supervision to encourage plausible dynamics. PhysPT~\cite{zhang2024physpt} proposes a neural module that refines kinematic motion using differentiable Euler-Lagrange losses to enforce rigid-body dynamics. IPMAN~\cite{tripathi2023intuitivephysics} incorporates intuitive physics cues through loss functions into monocular pose estimation, but remains a kinematics-based approach without enforcing full physical dynamics.
D\&D~\cite{li2022dnd} refines kinematic motion by estimating external forces and applying analytical physical computation to enforce consistency with Newtonian dynamics. 
While these methods improve physical realism to some extent, they operate as post-hoc refinement on kinematic reconstructions, making it difficult to recover from the ambiguity in the kinematics-based human mesh recovery stage. Moreover, the physical consistency is enforced through neural approximations rather than explicit physical simulation, leaving the overall pipeline fundamentally kinematics-based and decoupled from physical control. 

\subsection{Physics-based Human Motion Imitation}
Physics simulation platforms~\cite{makoviychuk2021isaac, todorov2012mujoco}, combined with reinforcement learning, have enabled physically grounded control of simulated characters, producing highly realistic human motion~\cite{2018-TOG-deepMimic, wang2024skillmimic, AMP, 2022-TOG-ASE, dou2022case, tessler2023calm}. 
PPR~\cite{yang2023ppr} leverages physics priors for plausible video-based reconstruction, and differentiable dynamics models~\cite{gartner2022diffphy} integrate physics into end-to-end optimization. By training policies on large-scale motion capture datasets~\cite{AMASS:ICCV:2019, peng2021neural, kobayashi2023motion}, many works have demonstrated high-fidelity motion imitation through learned control policies~\cite{wagener2022mocapact, Luo2022EmbodiedSH, luo2024universal, Luo2023PerpetualHC, tessler2024maskedmimic, peng2018sfv, winkler2022questsim}. PhysCap~\cite{shimada2020physcap} constrains monocular capture with real-time physical simulation. However, these policies are trained to track clean 3D motion references, struggling to generalize when such data is unavailable. PHC~\cite{Luo2023PerpetualHC} estimates 3D keypoints from video as motion references, but its two-stage design decouples control from visual input, often leading to jitter and unnatural motion.

Moreover, prior methods rely heavily on reinforcement learning, which typically suffers from low sample efficiency. Hence, they struggle to fully exploit rich visual information and instead depend primarily on sparse, deterministic inputs such as 3D keypoints or kinematics-based representations. 
simXR~\cite{Luo_2024_CVPR} employs a distillation-only scheme in a VR setting to train vision-to-action policies. While this avoids the need for reinforcement learning, it lacks robustness due to limited data and the absence of exploration. In contrast, our joint PPO+Distillation training substantially improves stability and generalization, demonstrating clear advantages over a pure distillation approach.

Learning vision-conditioned policies for human motion reconstruction that directly aligns with visual evidence remains a largely underexplored challenge.

\section{Preliminaries}

We formulate physically plausible human motion reconstruction as a goal-conditioned, physics-based motion imitation problem. Specifically, we use deep reinforcement learning (DRL) to train a policy that drives a simulated humanoid \cite{Luo2023PerpetualHC} to imitate motion sequences within a physical environment, with the goal signals as guidance. The policy, $\pi$, is modeled as a Markov Decision Process (MDP), defined by the tuple
\(\mathcal{M} = \langle S, A, T, R, \gamma \rangle\), where \( S \), \( A \), \( T \), \( R \), and \( \gamma \) denote the state space, action space, transition dynamics, reward function, and discount factor, respectively.

At each timestep \(t\), the state \(s_t\) consists of proprioceptive information \(s_t^p\) and goal information \(s_t^g\). Here, \(s_t^p\) includes the local 3D pose \(q_t\) and velocity \(\dot{q}_t\). In traditional motion imitation tasks, the goal \(s_t^g\) is typically defined by a reference trajectory \((\theta_t, \Gamma_t, \tau_t)\), encoding the local pose, global translation, and global rotation. In our method, the goal information is extracted from the input video, including frame-level visual features
and global spatial guidance computed by a pixel-as-ray strategy. This design enables the policy to directly leverage visual observations for motion imitation (see Sec.~\ref{sec:method}).

The action, \( a_t \), specifies target joint rotations, which are provided as control targets to a proportional-derivative 
(PD) controller to generate physically valid motion. 
 At each timestep \( t \), the humanoid agent samples an action \( a_t \in A\) from the policy \( \pi(a_t | s_t)\), 
where \( s_t \in S\) is the current state of the humanoid. The action is then executed in a physics simulator, 
producing the next state \( s_{t+1} = T(s_t, a_t)\) and the reward $r_t = R(s_t, a_t)$ for this action. We optimize the policy using Proximal Policy Optimization (PPO), with the objective of maximizing the expected discounted return:
\(\mathbb{E} \left[ \sum_{t=1}^{N} \gamma^{t-1} r_t \right].
\)

\section{Method}
\label{sec:method}

Fig. \ref{fig:pipeline} provides an overview of our method.
Given a monocular video with $N$ frames $\{I_t\}_{t=1}^{N}$, our goal is to reconstruct a physically plausible human motion sequence, which consists of local poses $\{\theta_t \in \mathbb{R}^{23\times 3} \}_{t=1}^{N}$, global translation $\{\Gamma_t \in \mathbb{R}^3 \}_{t=1}^{N}$, and orientation $\{\tau_t \in \mathbb{R}^3 \}_{t=1}^{N}$ in the world. 


Our visual-to-action policy, PhysHMR, starts by extracting local visual features for each video frame using a pre-trained HMR model~\citep{shen2024gvhmr}. These features are then used as input to the policy network, which predicts control actions for the humanoid (Sec. \ref{4.1}). 
To provide spatial grounding, we propose a pixel-as-ray strategy that lifts 2D keypoints into 3D rays and transforms them into global space using camera poses.
These global rays provide soft global guidance to the policy without enforcing strict positional constraints (Sec. \ref{4.2}). 
Finally, we enhance the training of our visual-to-action policy through knowledge distillation from a pre-trained motion imitation expert, leading to improved sample efficiency and policy robustness (Sec. \ref{4.3}).



\subsection{Local Reference from Visual Observations}
\label{4.1}

Prior works~\cite{shen2024gvhmr,ye2023slahmr} show that local motion features that capture relative joint articulation while being invariant to global root transformations are critical for effective motion learning. 
Such features are difficult to infer explicitly from images due to camera motion and depth ambiguity. 
Hence, we propose to use the pretrained video encoder from GVHMR~\cite{shen2024gvhmr}, which is trained to predict SMPL joint rotations relative to their parent joints. This naturally yields root-invariant visual features that serve as structured and physically meaningful input for local control. Unlike explicit pose reconstructions that commit to a single, potentially inaccurate estimate, these visual features retain rich pose-related information without collapsing to a deterministic pose.

Given video frames $\{I_t\}_{t=1}^{N}$, we first preprocess each frame $I_{t}$ to extract image features~\cite{goel2023humans}, bounding boxes~\cite{yolov8, li2022cliff}, 2D keypoints~\cite{xu2022vitpose}, and relative camera rotations~\cite{NEURIPS2023_7ac484b0}, denoted as $f^{\text{feat}}_t,f^{\text{bbox}}_t,f^{\text{kp2d}}_t,f^{\text{cam}}_t$, respectively. These per-frame features are then fed into the video encoder, which aggregates information across frames:
\[
\{F_t\}_{t=1}^{N} = \text{Enc}_{\text{GVHMR}}(\{f^{\text{feat}}_t,f^{\text{bbox}}_t,f^{\text{kp2d}}_t,f^{\text{cam}}_t\}_{t=1}^{N}) \in \mathbb{R}^{N \times D},
\]
where \(D\) is the feature dimension. The cross-frame fusion process not only enhances stability under occlusion but also supports flexible feature masking, allowing the model to operate with partial inputs. This flexibility enables the use of motion-only datasets like AMASS~\cite{AMASS:ICCV:2019} during training, even without paired RGB images, i.e., $f^{\text{feat}}_t$ is dropped.

Although local features bolster reconstruction robustness, accurate physics learning still demands global guidance, because simulations must be carried out in the world coordinate frame.
Thus, we enhance the local observation by leveraging GVHMR’s multitask MLP head to explicitly regress the future root orientation \(\bar{\tau}_{t+1}\) from the visual features \(F_t\). This auxiliary prediction provides a forward-looking estimate of the global root orientation in the camera coordinate system. We transform \(\bar{\tau}_{t+1}\) into the world frame and compute its relative difference from the current root pose \(\tau_t\) of the humanoid agent as:
\[
\Delta \tau_t = \tau_t^{-1} \bar{\tau}_{t+1}.
\]
This signal provides an explicit orientation cue that guides the agent’s future heading. We include both the visual feature \(F_t\) and the relative root orientation \(\Delta \tau_t\) in the observation passed to the policy at each timestep $t$.

\subsection{Global Guidance via Pixel-as-Ray}
\label{4.2}

Accurate global positioning is critical for physically plausible motion reconstruction, especially when camera motion is involved. However, directly predicting 3D trajectories from a monocular video is often unreliable due to depth ambiguity and motion noise. These trajectory errors can significantly degrade the performance of tracking-based control policies, leading to unstable motion. To circumvent this issue, we propose a pixel-as-ray strategy that encodes global guidance without enforcing explicit positional targets.

\subsubsection{Keypoint Lifting to 3D Rays}
Given extracted 2D keypoints, \(f_{t}^{\text{kp2d}} = \{(u^i_t, v^i_t)\}_{i=1}^{J}\), that represent the image-space locations of each joint \(i\) of the simulated humanoid in frame \(t\), and the camera intrinsics matrix \(\mathbf{K}\), we back-project each keypoint to obtain a 3D ray in the camera coordinate system:
\begin{equation}
    \text{ray}^i_t(s) = \mathbf{o}_t + s \cdot \mathbf{r}^i_t, \quad s > 0,
\end{equation}
\begin{equation}
o_t = \begin{bmatrix}
    0\\
    0\\
    0
\end{bmatrix}, \quad
    \mathbf{r}^i_t = \mathbf{K}^{-1}
\begin{bmatrix}
u^i_t \\
v^i_t \\
1
\end{bmatrix},
\end{equation}
where $\mathbf{o}_t$ is the camera origin in frame $t$, and $\mathbf{r}^i_t$ is the viewing direction for keypoint $i$.
This ray represents all possible 3D positions of joint $i$ along the corresponding viewing direction.

To align the ray with the simulation world, we further transform it using the camera-to-world transformation  $\mathbf{T}_t^{\text{c2w}}$, which is estimated via off-the-shelf methods~\cite{shen2024gvhmr,TRAM}:

\begin{equation}
\hat{\mathbf{o}}_t = \mathbf{T}_t^{\text{c2w}} \cdot h(\mathbf{o}_t), \quad 
\hat{\mathbf{r}}^i_{t} = \mathbf{T}_t^{\text{c2w}} \cdot h(\mathbf{r}^i_t), 
\end{equation}
where $\mathbf{T}_t^{\text{c2w}}\in \mathbb{R}^{3\times4}$ is a transformation matrix consisting of rotation and translation, $h(\cdot)$ is the homogeneous lifting function that augments a 3D vector with a $1$, $\hat{\mathbf{o}}_t$ denotes the ray origin in world coordinates, and $\hat{\mathbf{r}}^i_{t}$ is the transformed ray direction for keypoint $i$.

\subsubsection{Computing Ray Displacement Vectors}
For each humanoid joint $i$ at time $t$, we compute the shortest vector from the joint position $\mathbf{j}^i_t$ to the corresponding ray defined by origin $\hat{\mathbf{o}}_t$ and direction $\hat{\mathbf{r}}^i_t$, yielding a displacement vector $\mathbf{d}^i_t$:
\begin{equation}
    \mathbf{d}^i_t = \text{proj}_{\perp}(\mathbf{j}^i_t, \hat{\mathbf{o}}_t, \hat{\mathbf{r}}^i_t),
\end{equation}

\noindent where $\text{proj}_{\perp}(\cdot)$ denotes the perpendicular offset from the point $\mathbf{j}^i_t$ to the ray \(\hat{\mathbf{o}}_t + s \cdot \hat{\mathbf{r}}^i_t\), and $\mathbf{j}^i_t$ is obtained from the simulated humanoid proprioception $\mathbf{s}_t^p$.
These displacements $\{\mathbf{d}^i_t\}_{i=1}^J$ are concatenated and passed to the policy network as global spatial observations. Compared to using noisy 3D joint positions, this formulation enables more flexible and robust spatial grounding for humanoid control.

Unlike reprojection error, which is typically used as a training loss, our pixel-as-ray formulation is explicitly used as part of the policy input, allowing the network to exploit these signals during inference for robust global alignment.

To account for potentially unreliable 2D keypoint estimates, 
we append the keypoint confidence scores predicted by the 2D keypoint estimator to the displacement vectors, enabling the policy to adaptively modulate its reliance on uncertain inputs. Additionally, inspired by \cite{goel2023humans}, we introduce random masking and perturbation of keypoint inputs during training to improve robustness under in-the-wild conditions.

\subsection{Policy Learning with Reinforcement and Distillation}
\label{4.3}

\subsubsection{Distillation} Although we've used a pretrained visual encoder to extract informative features from monocular images, directly training a control policy from these features using reinforcement learning remains highly sample-inefficient. To address this, we introduce a knowledge distillation framework that transfers motion expertise from a pretrained teacher policy, \(\pi_{\text{teach}}(a_t\mid s_t^p, \theta_t)\), which is trained to perform standard motion imitation using ground-truth pose supervision from the AMASS dataset~\cite{AMASS:ICCV:2019}. The teacher policy takes as input the agent's proprioceptive state \(s_t^p\) and the target kinematic pose \(\theta_t\), and outputs physically valid actions that track the reference motion.

Given paired supervision data \((I_t, \theta_t)\), where \(I_t\) is the input image and \(\theta_t\) is the corresponding target kinematic pose, we use the teacher's action as a supervision signal to guide the training of our visual-to-action policy, \(\pi_{\text{PhysHMR}}(a_t \mid s_t^p, F_t, \Delta \tau_t, \{\mathbf{d}_t^i\}_{i=1}^{J})\). This policy takes as input the agent’s proprioceptive state \(s_t^p\), frame-level visual features \(F_t\), relative root orientation \(\Delta \tau_t\), and pixel-to-ray spatial displacements \(\{\mathbf{d}_t^i\}_{i=1}^{J}\), and is trained to imitate the teacher’s actions without using ground-truth pose targets as explicit input. The training objective minimizes a distillation loss between the actions predicted by the student and teacher policies:
\[
L_{\text{distill}} = \left\| \pi_{\text{PhysHMR}}(a_t \mid s_t^p, F_t, \Delta \tau_t, \{\mathbf{d}_t^i\}_i) - \pi_{\text{teach}}(a_t \mid s_t^p, \theta_t) \right\|^2
\]

\noindent Note that both policies operate on the same proprioceptive state \(s_t^p\) obtained from simulation, but differ in how goal information is provided: the teacher policy is conditioned on the explicit ground-truth pose \(\theta_t\), while the PhysHMR policy relies on features extracted from the input image.
This objective encourages our PhysHMR policy to learn a control strategy capable of effectively imitating motion depicted in the input video without explicit pose signals.

\subsubsection{Overall Loss}
While distillation enables efficient learning from a strong teacher, it is inherently limited by the accuracy and diversity of the teacher’s actions. To enable more accurate and adaptable motion control that goes beyond the fixed supervision provided by the teacher, we complement supervised knowledge distillation with reinforcement learning, allowing the PhysHMR policy to refine its behavior through dynamic interaction with the environment. Specifically, we utilize a composite reward function:
\begin{equation}
R(s_t^{p}, \theta_t) = \alpha_1 R_{\text{pose}} + \alpha_2 R_{\text{amp}} + \alpha_3 R_{\text{energy}},
\label{eq:reward}
\end{equation}
where \(\theta_t\) is the target reference pose. \(R_{\text{pose}}\) is an imitation reward that promotes alignment with the reference pose, \(\theta_t\) \cite{Luo2023PerpetualHC}. \(R_{\text{amp}}\) is a style-based reward via Adversarial Motion Priors (AMP)~\cite{AMP} 
that encourages the generation of realistic, human-like motion aligned with the motion prior. \(R_{\text{energy}}\) is an energy reward that penalizes excessive joint accelerations to improve smoothness and reduce jitter~\cite{10.1145/3550469.3555411}.  Additional details about the reward formulation are provided in the suppl. document.

The overall training objective combines both supervised and reinforcement signals:
\begin{equation}
L = L_{\text{distill}} + L_{\text{PPO}},
\end{equation}
where \(L_{\text{PPO}}\) is the loss term calculated by PPO using the reward in Eq.~\ref{eq:reward}. By jointly training with distillation and reinforcement learning, our visual-to-action policy benefits from both sample-efficient supervision and environment-driven refinement, resulting in not only accelerated convergence but also enhanced policy performance.

\section{Experiments}
\begin{figure*}[t]
    \centering
    \includegraphics[width=0.98\linewidth]{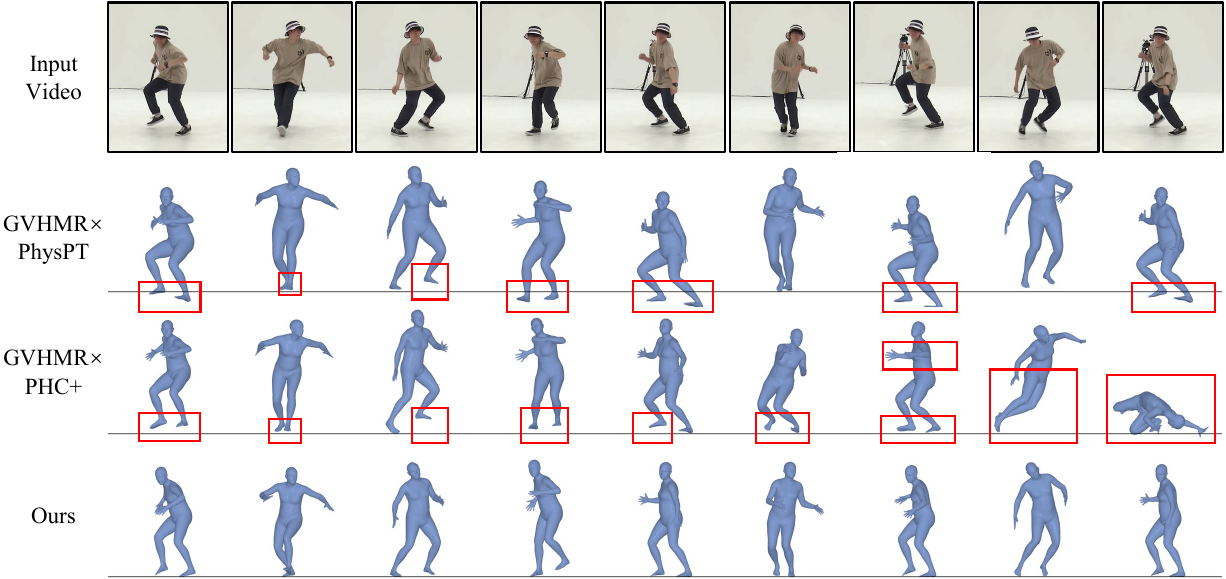}
\caption{
Comparison against two physics-based methods. The black line indicates the ground. PhysPT (row 2) uses neural networks to approximate physics, but still suffers from ground penetration. PHC+ (row 3) amplifies motion reconstruction errors during tracking, leading to unstable results. Both methods cannot correct upstream errors. In contrast, our visual-to-action approach produces motion that is both physically plausible and visually aligned.
}
  \Description{Comparison}
    \label{results}
\end{figure*}

\begin{table*}[t]
\centering
\caption{Comparison of our motion reconstruction variants on AIST++ and EMDB2 under kinematic and physical plausibility metrics. Lower is better.}
\label{tab:main}
\small
\begin{tabular}{l l 
                c c c c c c c 
                c c c c c c c}
\toprule
\multirow{2}{*}{{Phys. Type}} & \multirow{2}{*}{{Method}} 
& \multicolumn{7}{c}{{EMDB2}} 
& \multicolumn{7}{c}{{AIST++}} \\
\cmidrule(lr){3-9} \cmidrule(lr){10-16}
 &  & PA & WA & MPJ & FS & HV & ACC & VEL
     & PA & WA & MPJ & FS & HV & ACC & VEL \\
\midrule
\multirow{2}{*}{Kin.} 
    & TRAM               &  \textbf{35.51}  &  \textbf{148.05}  & \underline{56.74}  &  11.76 & 22.97  &  \textbf{4.77}  & \textbf{8.77 }  
                          & \textbf{50.18}  & 189.44 &  \underline{76.30} &  23.18 &  7.93 &  9.31 &   21.85 \\
    & GVHMR             &  40.95 &  228.67  &  65.21 &  \underline{5.65}  & 26.42   &  5.40 &  \underline{10.19}
                        &  53.43 &  \textbf{175.64}  &  79.05 &  11.54  &  4.64   &  10.19 & \underline{14.40} \\
\midrule
\multirow{3}{*}{Neural} 
    & PhysPT (CLIFF)        & 48.40  &  762.78   &  77.00  &  11.02 &  6.54 & 6.72  &  19.92 
                          & 70.72  &  260.07  &  108.57  &  13.68  &  3.71  &  9.21  &  17.05 \\
    & TRAM $\times$ PhysPT       &   39.90  & 704.57  &  61.42  &  8.49  &  7.02  &   5.38 &  17.71
                          &   52.79  &  250.30   &  83.93  &  \underline{10.94} &  3.55  & \underline{8.59}  &  16.00  \\
    &  GVHMR $\times$ PhysPT    &  41.34  &   682.03  &  66.08   &  10.71 &  8.46   &  \underline{5.35 }  & 17.24 
                          & 55.29   & 235.66   &  83.93  &  11.24 &  \underline{3.46}  &  8.90 &  15.36  \\
\midrule
\multirow{2}{*}{Track.} 
    & TRAM $\times$ PHC+           & 52.94 & \underline{158.58} & 74.34  & 23.41 & 7.64 & 9.56  & 14.00 
                         & 71.21 &  212.28 & 101.70  & 36.57  & 4.95  & 12.55   &  22.74     \\
    & GVHMR $\times$ PHC+          &  46.24 & 193.01 &  72.50  & 12.71 &  7.71 & 7.43  &  12.21
                          & 67.38 & 193.23 & 109.77  & 24.79  &  6.05 & 10.05  & 17.10   \\
\midrule

V2A 

    & \textbf{Ours} & \underline{39.34}  &   189.26  &  \textbf{55.48}  &   \textbf{4.60}  & \textbf{5.04} & {5.49}   &  10.53     
                  &   \underline{50.40}  &  \underline{187.42}  &  \textbf{63.94}  &   \textbf{9.14} &  \textbf{3.10} &  \textbf{6.58}  & \textbf{12.13}   \\
\bottomrule
\end{tabular}
\end{table*}

\subsection{Implementation Details}
\label{sec:implementation}

We use Isaac Gym~\cite{makoviychuk2021isaac} as the physics simulator and train our model on a single NVIDIA L40 GPU. The physics simulation runs at 60 Hz, while control actions are issued at 30 Hz. All videos and motion sequences are sampled at 30 FPS for consistency. The physical model parameters (e.g., masses, joint torque limits, friction coefficients) all follow the settings in PHC~\cite{Luo2023PerpetualHC}.

We parallelize training with 1,536 environments to improve sample efficiency. The main policy network is implemented as an MLP with hidden layer dimensions of [2048, 1536, 1024, 1024, 512, 512] and SiLU as the activation function. Reinforcement learning is conducted with Proximal Policy Optimization (PPO), using a clip coefficient of 0.2. PHC+\cite{luo2024universal}
is used as the teacher policy. The distillation loss is jointly optimized with the PPO objective. 
We apply gradient clipping with a threshold of 50 to ensure stability. Early termination is enabled to reduce ineffective exploration on failed episodes and accelerate convergence. The model typically converges after approximately three days of training. 

The current pipeline depends on the pretrained GVHMR image encoder, which is not real-time, and therefore the entire system operates offline. Extending this framework with causal attention and efficient encoders could make online deployment feasible in future work.

Our approach does not rely on explicit shape information. We estimate the human shape parameters of the SMPL model with an off-the-shelf tool and compute the scale difference relative to a zero-shape SMPL model. This scales the simulation space to match real-world units, such that the humanoid retains canonical zero-shape. 

\subsection{Datasets}
We utilize Human3.6M \cite{h36m_pami},  AIST++ \cite{li2021learn}, EMDB2 \cite{kaufmann2023emdb}, and AMASS \cite{AMASS:ICCV:2019} for our experiments.
Human3.6M contains 3.6M 3D human poses from 11 actors across 4 viewpoints, with accurate 3D keypoints but no SMPL ground truth. We exclude sequences involving chairs to avoid simulator inconsistencies. SMPL parameters are estimated via GVHMR and refined using LBFGS by aligning SMPL joints with the 3D keypoints. Note that since the full Human3.6M dataset is used in training of both TRAM and GVHMR, we only use it to conduct ablation studies.
We use a zero-shape SMPL model and introduce an additional scale parameter to account for individual body proportions~\ref{sec:implementation}.
AIST++ provides 1,408 dance sequences from 30 subjects across 9 views, featuring dynamic and diverse movements that are challenging for physics-based motion learning. AIST++ provides only a scale parameter without explicit shape information. AMASS is a large-scale, image-free motion capture dataset. EMDB2 contains long-range, moving-camera sequences. We remove sequences like skateboarding and stair climbing that are incompatible with simulation. We train PhysHMR on the combined training splits of Human3.6M, AIST++, and AMASS (image-free). Only AMASS is used for the AMP reward. EMDB2 is used for evaluation only.

\subsection{Metrics}  
To evaluate the accuracy and physical plausibility of the reconstructed motion, we use the following metrics:
(1) \textbf{MPJ} (Mean Per Joint Position Error, MPJPE, mm): Measures the average Euclidean distance between predicted and ground-truth 3D joint positions after aligning the root joint. 
(2) \textbf{WA} (World-aware MPJPE, WA-MPJPE, mm): Similar to MPJPE, but computed in the global coordinate system, capturing errors in both pose and global translation.
(3) \textbf{PA} (Procrustes Aligned MPJPE, PA-MPJPE, mm): Computes MPJPE after applying rigid alignment (scale, rotation, translation) to isolate pose errors independent of global position.
(4) \textbf{VEL} (Velocity Error, mm/s) and (5) \textbf{ACC} (Acceleration Error, mm/s\(^2\)): Measure the temporal consistency of joint movement across frames.

To assess physical realism, we introduce a new metric: (6) \textbf{HV} (Foot Height Variance, mm): In every frame, we record the vertical position of the lowest foot joint. We select the lowest 25 \% across all frames and compute their variance; smaller HV indicates more stable, physically realistic contact. This kinematic proxy evaluates contact consistency without requiring an explicit ground plane.
Additionally, we use (7) \textbf{FS} (Foot Sliding, mm) to measure undesired foot movement when the foot is expected to be in contact with the ground.
Together, these metrics provide a comprehensive evaluation of both motion accuracy and physical plausibility.

Physics-based methods are less stable than kinematic ones: once a failure occurs, the humanoid usually falls and remains collapsed, causing large errors to dominate the averages. To mitigate this, we split all test sequences into 100-frame clips and evaluate them individually. This protocol, also common in kinematics-based methods, ensures fairness. Following PHC+, we compute metrics only on successful clips (discarding those with PA-MPJPE > 100), which avoids excessive sequence removal while keeping the results representative.

\subsection{Comparisons}  

We compare our method with both kinematic and physics-based state-of-the-art approaches. Kinematic methods, TRAM~\cite{TRAM} and GVHMR~\cite{shen2024gvhmr}, estimate human motion from videos without enforcing physical constraints. In contrast, PhysPT~\cite{zhang2024physpt} introduces a physics-based approach by first estimating SMPL parameters using CLIFF~\cite{li2022cliff} and then refining the motion with a transformer-based model to improve physical plausibility. We also provide results for GVHMR $\times$ PhysPT and TRAM $\times$ PhysPT, where the SMPL estimation backbone of PhysPT is replaced with TRAM and GVHMR, respectively, to ensure a fair comparison. 
Additionally, we evaluate tracking-based methods, TRAM $\times$ PHC+ and GVHMR $\times$ PHC+, where the tracking policy PHC+~\cite{luo2024universal} is applied to track the outputs of TRAM and GVHMR, respectively, providing a direct comparison between motion reconstruction via traditional tracking policies and our visual-to-action policy.

For fair comparison, all baselines rely on global human trajectories: GVHMR and TRAM each estimate their own, and variants (e.g., GVHMR × PHC+, TRAM × PHC+) follow them. Our method instead leverages the camera trajectory and 2D keypoints to form the pixel-as-ray input. Since TRAM estimates extrinsics while GVHMR does not, we use TRAM’s camera trajectory, rigidly aligned to GVHMR’s first-frame coordinates, to provide consistent camera input. 

We use the GVHMR encoder for image features, ensuring fairness on the vision side. For physics, policies trained on high-dynamic datasets (e.g., AIST++) are overly sensitive to noisy estimates, so we adopt PHC+ as the tracker baseline.

\subsubsection{Quantitative Results}
As shown in Tab. \ref{tab:main}, kinematic-based methods generally achieve lower errors on MPJPE, PA-MPJPE, and WA-MPJPE, as they are directly optimized to minimize 3D keypoint discrepancies and ignore physical constraints. In contrast, physics-based approaches trade off keypoint accuracy for physical plausibility. For example, PhysPT, TRAM $\times$ PhysPT, and GVHMR $\times$ PhysPT all achieve better physical metrics due to PhysPT's physics-aware design. However, its global trajectory relies on foot-ground contact prediction, which can be inaccurate and result in high WA-MPJPE, especially for long-range motions in EMDB2. 

Traditional tracking-based methods, TRAM $\times$ PHC+ and GVHMR $\times$ PHC+, exhibit stable performance when the kinematic estimates are accurate, but their quality degrades severely when those estimates are poor, as seen on AIST++, where the challenging motions lead to high PA-MPJPE. Moreover, such methods fails to capitalize on physical simulation, with subpar FS and HV scores. This is due to excessive movements of the limbs during balance recovery, which degrades physical metrics.

In contrast, our method achieves state-of-the-art performance on FS and HV, demonstrating superior physical realism. It also remains competitive across MPJPE metrics. By learning policies directly from visual features, our approach can produce 3D human motion that is both physically plausible and visually aligned with the input video.

\subsubsection{User Study}
To further evaluate perceptual quality beyond quantitative error metrics, we conducted a user study comparing PhysHMR against PhysPT and GVHMR~$\times$~PHC+. 
Participants (26 in total) were presented with 5 groups of side-by-side videos and asked to select the result they perceived as more visually aligned with the input video and physically plausible. Overall, 66.3\% of participants preferred PhysHMR, compared to 19.9\% for PhysPT and 13.8\% for GVHMR~$\times$~PHC+ (see Table~\ref{tab:user-study}), indicating that our method not only improves numerical accuracy but also provides higher perceptual fidelity. 

\begin{table}[h]
\centering
\caption{User study preference results. Values indicate the percentage of times each method was preferred.}
\label{tab:user-study}
\begin{tabular}{lccc}
\toprule
Method & PhysHMR  & PhysPT & GVHMR~$\times$~PHC+ \\
\midrule
Preference (\%) & \textbf{66.3} & 19.9 & 13.8 \\
\bottomrule
\end{tabular}
\end{table}

\subsection{Ablation Study}  
We conduct ablation experiments on H36M to validate the effectiveness of our proposed \textit{pixel-as-ray} formulation and the combined training strategy based on distillation and reinforcement learning.

\noindent \textbf{Effect of Pixel-as-Ray.}  
Table~\ref{tab:ablation-obs} evaluates the impact of different global instruction strategies. Removing the global instruction entirely (ImgFeat) yields good PA-MPJPE and MPJPE, but significantly worse WA-MPJPE, indicating that the humanoid mimics local motions well but fails to track global trajectories. Replacing pixel-as-ray with global supervision from explicit root-relative displacements estimated with GVHMR (+ 3D root) results in degraded performance across all metrics, as errors in root estimation introduce misleading guidance that conflicts with local motion. In contrast, using 2D keypoints via pixel-as-ray (+ pixelray) provides more robust and relaxed global instruction, achieving comparable PA-MPJPE to the no-global setting while substantially improving WA-MPJPE.

\begin{figure}[t]
    \centering
    \includegraphics[width=0.98\linewidth]{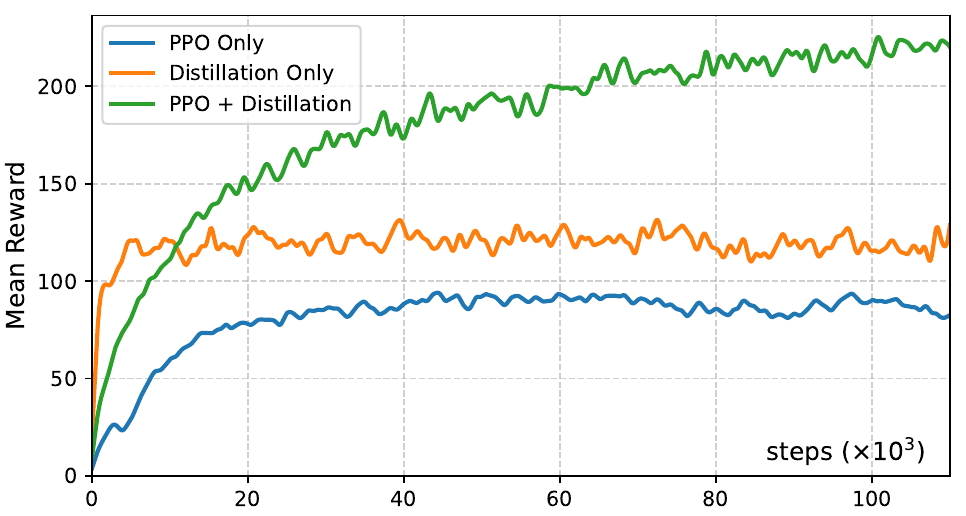}
\caption{
Mean reward curves during training. PPO Only converges slowly and underperforms. Distillation Only converges quickly but plateaus early. Our approach (PPO + Distillation) achieves both faster convergence and higher final rewards.
}
  \Description{reward curves}
    \label{fig:training-curves}
\end{figure}

\noindent \textbf{Effect of Distillation.}  
Table~\ref{tab:ablation-policy} and Figure~\ref{fig:training-curves} compare different training strategies. We also report success rates, defined as the percentage of sequences where PA-MPJPE remains below 50 mm for all frames. Combining PPO with distillation achieves the highest success rate, showing that PPO substantially improves long-term stability.
Using PPO Only leads to slow convergence and suboptimal final performance. 
The distillation-only setting enables faster early-stage learning but lacks exploration, resulting in limited reward improvements. Note that in the Distillation Only setting, the reward is computed only for evaluation and not used during training. Our joint training strategy combines the strengths of both: it accelerates convergence and achieves higher final rewards, while also delivering better generalization on test sequences.

\begin{table}[t]
  \centering
  \caption{Ablation on global instruction strategies. }
  \label{tab:ablation-obs}
  \begin{tabular}{lccc}
    \toprule
    Obs. Type   & PA ↓  & WA ↓ &   MPJ ↓  \\
    \midrule
    ImgFeat     &   \textbf{35.85}     &   142.17            &    47.78   \\
    + 3D root      &   38.70              &   195.59            &    65.03   \\
    + pixelray (Ours)     &   36.05              &   \textbf{112.60}   &    \textbf{47.01}   \\
    \bottomrule
  \end{tabular}
\end{table}

\begin{table}[t]
  \centering
  \caption{Comparison of policy learning strategies. Combining reinforcement learning (PPO) and distillation yields the best performance.}
  \label{tab:ablation-policy}
  \begin{tabular}{lcccc}
    \toprule
    Strategy & PA ↓ & WA ↓ & MPJ ↓ &SR ↑ \\
    \midrule
    PPO Only       &    42.18    &  117.33  &    58.25 &  65.5\%       \\
    Distill. Only       &    39.62    &  114.69  &    52.41  & 72.0\%   \\
    PPO + Distill.       &  \textbf{36.05}  &  \textbf{112.60}  &  \textbf{47.01}  & \textbf{88.4\%}\\
    \bottomrule
  \end{tabular}
\end{table}

\section{Conclusion}
We presented PhysHMR, a unified framework for reconstructing physically plausible human motion from monocular videos by directly mapping visual inputs to humanoid control actions. Unlike prior methods, PhysHMR learns a visual-to-action policy that integrates physical dynamics during inference. To improve efficiency and robustness, we introduce motion distillation from a mocap-trained expert and a novel pixel-as-ray strategy that provides soft global guidance without relying on noisy 3D root predictions. 

\noindent \textbf{Limitation and Future Work.}
While PhysHMR generates high-fidelity motion, a real-to-sim gap persists due to differences in body mechanics and contact properties, which can sometimes lead to visible artifacts.  Future work will incorporate personalized physical parameters to better reflect real-world dynamics. Additionally, motion reconstruction from a single monocular video is underconstrained due to ambiguity and occlusion; using a conditional generative model instead of a deterministic policy may better capture diverse and physically plausible motions. Our current framework does not explicitly support human–scene interactions (e.g., sitting or leaning against surfaces), which we plan to address through environment reconstruction and interaction-aware control in our future works.

\bibliographystyle{ACM-Reference-Format}
\bibliography{reference}


\begin{thebibliography}{55}


\ifx \showCODEN    \undefined \def \showCODEN     #1{\unskip}     \fi
\ifx \showISBNx    \undefined \def \showISBNx     #1{\unskip}     \fi
\ifx \showISBNxiii \undefined \def \showISBNxiii  #1{\unskip}     \fi
\ifx \showISSN     \undefined \def \showISSN      #1{\unskip}     \fi
\ifx \showLCCN     \undefined \def \showLCCN      #1{\unskip}     \fi
\ifx \shownote     \undefined \def \shownote      #1{#1}          \fi
\ifx \showarticletitle \undefined \def \showarticletitle #1{#1}   \fi
\ifx \showURL      \undefined \def \showURL       {\relax}        \fi
\providecommand\bibfield[2]{#2}
\providecommand\bibinfo[2]{#2}
\providecommand\natexlab[1]{#1}
\providecommand\showeprint[2][]{arXiv:#2}

\bibitem[Arnab et~al\mbox{.}(2019)]%
        {Arnab_CVPR_2019}
\bibfield{author}{\bibinfo{person}{Anurag Arnab}, \bibinfo{person}{Carl Doersch}, {and} \bibinfo{person}{Andrew Zisserman}.} \bibinfo{year}{2019}\natexlab{}.
\newblock \showarticletitle{Exploiting Temporal Context for 3D Human Pose Estimation in the Wild}. In \bibinfo{booktitle}{\emph{Proceedings of the IEEE/CVF Conference on Computer Vision and Pattern Recognition (CVPR)}}.
\newblock


\bibitem[Bogo et~al\mbox{.}(2016)]%
        {10.1007/978-3-319-46454-1_34}
\bibfield{author}{\bibinfo{person}{Federica Bogo}, \bibinfo{person}{Angjoo Kanazawa}, \bibinfo{person}{Christoph Lassner}, \bibinfo{person}{Peter Gehler}, \bibinfo{person}{Javier Romero}, {and} \bibinfo{person}{Michael~J. Black}.} \bibinfo{year}{2016}\natexlab{}.
\newblock \showarticletitle{Keep It SMPL: Automatic Estimation of 3D Human Pose and Shape from a Single Image}. In \bibinfo{booktitle}{\emph{Proceedings of the European Conference on Computer Vision (ECCV)}}, \bibfield{editor}{\bibinfo{person}{Bastian Leibe}, \bibinfo{person}{Jiri Matas}, \bibinfo{person}{Nicu Sebe}, {and} \bibinfo{person}{Max Welling}} (Eds.).
\newblock


\bibitem[Cai et~al\mbox{.}(2023)]%
        {cai2023smplerx}
\bibfield{author}{\bibinfo{person}{Zhongang Cai}, \bibinfo{person}{Wanqi Yin}, \bibinfo{person}{Ailing Zeng}, \bibinfo{person}{Chen Wei}, \bibinfo{person}{Qingping Sun}, \bibinfo{person}{Yanjun Wang}, \bibinfo{person}{Hui~En Pang}, \bibinfo{person}{Haiyi Mei}, \bibinfo{person}{Mingyuan Zhang}, \bibinfo{person}{Lei Zhang}, \bibinfo{person}{Chen~Change Loy}, \bibinfo{person}{Lei Yang}, {and} \bibinfo{person}{Ziwei Liu}.} \bibinfo{year}{2023}\natexlab{}.
\newblock \showarticletitle{{SMPLer-X}: Scaling Up Expressive Human Pose and Shape Estimation}. In \bibinfo{booktitle}{\emph{Advances in Neural Information Processing Systems (NeurIPS), Datasets and Benchmarks Track}}.
\newblock


\bibitem[Dou et~al\mbox{.}(2023)]%
        {dou2022case}
\bibfield{author}{\bibinfo{person}{Zhiyang Dou}, \bibinfo{person}{Xuelin Chen}, \bibinfo{person}{Qingnan Fan}, \bibinfo{person}{Taku Komura}, {and} \bibinfo{person}{Wenping Wang}.} \bibinfo{year}{2023}\natexlab{}.
\newblock \showarticletitle{C·ASE: Learning Conditional Adversarial Skill Embeddings for Physics-based Characters}. In \bibinfo{booktitle}{\emph{SIGGRAPH Asia Conference Papers (SA Conference Papers)}}.
\newblock


\bibitem[G{\"a}rtner et~al\mbox{.}(2022)]%
        {gartner2022diffphy}
\bibfield{author}{\bibinfo{person}{Erik G{\"a}rtner}, \bibinfo{person}{Mykhaylo Andriluka}, \bibinfo{person}{Erwin Coumans}, {and} \bibinfo{person}{Cristian Sminchisescu}.} \bibinfo{year}{2022}\natexlab{}.
\newblock \showarticletitle{Differentiable Dynamics for Articulated 3D Human Motion Reconstruction}. In \bibinfo{booktitle}{\emph{Proceedings of the IEEE/CVF Conference on Computer Vision and Pattern Recognition (CVPR)}}.
\newblock


\bibitem[Goel et~al\mbox{.}(2023)]%
        {goel2023humans}
\bibfield{author}{\bibinfo{person}{Shubham Goel}, \bibinfo{person}{Georgios Pavlakos}, \bibinfo{person}{Jathushan Rajasegaran}, \bibinfo{person}{Angjoo Kanazawa}, {and} \bibinfo{person}{Jitendra Malik}.} \bibinfo{year}{2023}\natexlab{}.
\newblock \showarticletitle{Humans in 4{D}: Reconstructing and Tracking Humans with Transformers}. In \bibinfo{booktitle}{\emph{Proceedings of the IEEE/CVF International Conference on Computer Vision (ICCV)}}.
\newblock


\bibitem[Huang et~al\mbox{.}(2017)]%
        {MuVS:3DV:2017}
\bibfield{author}{\bibinfo{person}{Yinghao Huang}, \bibinfo{person}{Federica Bogo}, \bibinfo{person}{Christoph Lassner}, \bibinfo{person}{Angjoo Kanazawa}, \bibinfo{person}{Peter~V. Gehler}, \bibinfo{person}{Javier Romero}, \bibinfo{person}{Ijaz Akhter}, {and} \bibinfo{person}{Michael~J. Black}.} \bibinfo{year}{2017}\natexlab{}.
\newblock \showarticletitle{Towards Accurate Marker-less Human Shape and Pose Estimation over Time}. In \bibinfo{booktitle}{\emph{Proceedings of the International Conference on 3D Vision (3DV)}}.
\newblock


\bibitem[Ionescu et~al\mbox{.}(2014)]%
        {h36m_pami}
\bibfield{author}{\bibinfo{person}{Catalin Ionescu}, \bibinfo{person}{Dragos Papava}, \bibinfo{person}{Vlad Olaru}, {and} \bibinfo{person}{Cristian Sminchisescu}.} \bibinfo{year}{2014}\natexlab{}.
\newblock \showarticletitle{Human3.6M: Large Scale Datasets and Predictive Methods for 3D Human Sensing in Natural Environments}.
\newblock \bibinfo{journal}{\emph{IEEE Transactions on Pattern Analysis and Machine Intelligence (TPAMI)}} (\bibinfo{year}{2014}).
\newblock


\bibitem[Jiang et~al\mbox{.}(2023)]%
        {jiang2023drop}
\bibfield{author}{\bibinfo{person}{Yifeng Jiang}, \bibinfo{person}{Jungdam Won}, \bibinfo{person}{Yuting Ye}, {and} \bibinfo{person}{C.~Karen Liu}.} \bibinfo{year}{2023}\natexlab{}.
\newblock \showarticletitle{DROP: Dynamics Responses from Human Motion Prior and Projective Dynamics}. In \bibinfo{booktitle}{\emph{SIGGRAPH Asia Conference Papers (SA Conference Papers)}}.
\newblock


\bibitem[Jocher et~al\mbox{.}(2023)]%
        {yolov8}
\bibfield{author}{\bibinfo{person}{Glenn Jocher}, \bibinfo{person}{Ayush Chaurasia}, {and} \bibinfo{person}{Jing Qiu}.} \bibinfo{year}{2023}\natexlab{}.
\newblock \bibinfo{title}{Ultralytics YOLOv8}.
\newblock \bibinfo{howpublished}{\url{https://github.com/ultralytics/ultralytics}}.
\newblock


\bibitem[Kaufmann et~al\mbox{.}(2023)]%
        {kaufmann2023emdb}
\bibfield{author}{\bibinfo{person}{Manuel Kaufmann}, \bibinfo{person}{Jie Song}, \bibinfo{person}{Chen Guo}, \bibinfo{person}{Kaiyue Shen}, \bibinfo{person}{Tianjian Jiang}, \bibinfo{person}{Chengcheng Tang}, \bibinfo{person}{Juan~Jos{\'e} Z{\'a}rate}, {and} \bibinfo{person}{Otmar Hilliges}.} \bibinfo{year}{2023}\natexlab{}.
\newblock \showarticletitle{{EMDB}: The Electromagnetic Database of Global 3D Human Pose and Shape in the Wild}. In \bibinfo{booktitle}{\emph{Proceedings of the IEEE/CVF International Conference on Computer Vision (ICCV)}}.
\newblock


\bibitem[Kobayashi et~al\mbox{.}(2023)]%
        {kobayashi2023motion}
\bibfield{author}{\bibinfo{person}{Makito Kobayashi}, \bibinfo{person}{Chen-Chieh Liao}, \bibinfo{person}{Keito Inoue}, \bibinfo{person}{Sentaro Yojima}, {and} \bibinfo{person}{Masafumi Takahashi}.} \bibinfo{year}{2023}\natexlab{}.
\newblock \bibinfo{title}{Motion Capture Dataset for Practical Use of AI-based Motion Editing and Stylization}.
\newblock
\showeprint[arxiv]{2306.08861}~[cs.CV]


\bibitem[Li et~al\mbox{.}(2022a)]%
        {li2022dnd}
\bibfield{author}{\bibinfo{person}{Jiefeng Li}, \bibinfo{person}{Siyuan Bian}, \bibinfo{person}{Chao Xu}, \bibinfo{person}{Gang Liu}, \bibinfo{person}{Gang Yu}, {and} \bibinfo{person}{Cewu Lu}.} \bibinfo{year}{2022}\natexlab{a}.
\newblock \showarticletitle{D\&D: Learning Human Dynamics from Dynamic Camera}. In \bibinfo{booktitle}{\emph{Proceedings of the European Conference on Computer Vision (ECCV)}}.
\newblock


\bibitem[Li et~al\mbox{.}(2021)]%
        {li2021learn}
\bibfield{author}{\bibinfo{person}{Ruilong Li}, \bibinfo{person}{Shan Yang}, \bibinfo{person}{David~A. Ross}, {and} \bibinfo{person}{Angjoo Kanazawa}.} \bibinfo{year}{2021}\natexlab{}.
\newblock \bibinfo{title}{Learn to Dance with AIST++: Music Conditioned 3D Dance Generation}.
\newblock
\showeprint[arxiv]{2101.08779}~[cs.CV]


\bibitem[Li et~al\mbox{.}(2022b)]%
        {li2022cliff}
\bibfield{author}{\bibinfo{person}{Zhihao Li}, \bibinfo{person}{Jianzhuang Liu}, \bibinfo{person}{Zhensong Zhang}, \bibinfo{person}{Songcen Xu}, {and} \bibinfo{person}{Youliang Yan}.} \bibinfo{year}{2022}\natexlab{b}.
\newblock \showarticletitle{CLIFF: Carrying Location Information in Full Frames into Human Pose and Shape Estimation}. In \bibinfo{booktitle}{\emph{Proceedings of the European Conference on Computer Vision (ECCV)}}.
\newblock


\bibitem[Loper et~al\mbox{.}(2015)]%
        {SMPL:2015}
\bibfield{author}{\bibinfo{person}{Matthew Loper}, \bibinfo{person}{Naureen Mahmood}, \bibinfo{person}{Javier Romero}, \bibinfo{person}{Gerard Pons-Moll}, {and} \bibinfo{person}{Michael~J. Black}.} \bibinfo{year}{2015}\natexlab{}.
\newblock \showarticletitle{{SMPL}: A Skinned Multi-Person Linear Model}.
\newblock \bibinfo{journal}{\emph{ACM Transactions on Graphics (TOG)}} (\bibinfo{year}{2015}).
\newblock


\bibitem[Luo et~al\mbox{.}(2024a)]%
        {Luo_2024_CVPR}
\bibfield{author}{\bibinfo{person}{Zhengyi Luo}, \bibinfo{person}{Jinkun Cao}, \bibinfo{person}{Rawal Khirodkar}, \bibinfo{person}{Alexander Winkler}, \bibinfo{person}{Kris Kitani}, {and} \bibinfo{person}{Weipeng Xu}.} \bibinfo{year}{2024}\natexlab{a}.
\newblock \showarticletitle{Real-Time Simulated Avatar from Head-Mounted Sensors}. In \bibinfo{booktitle}{\emph{Proceedings of the IEEE/CVF Conference on Computer Vision and Pattern Recognition (CVPR)}}.
\newblock


\bibitem[Luo et~al\mbox{.}(2024b)]%
        {luo2024universal}
\bibfield{author}{\bibinfo{person}{Zhengyi Luo}, \bibinfo{person}{Jinkun Cao}, \bibinfo{person}{Josh Merel}, \bibinfo{person}{Alexander Winkler}, \bibinfo{person}{Jing Huang}, \bibinfo{person}{Kris~M. Kitani}, {and} \bibinfo{person}{Weipeng Xu}.} \bibinfo{year}{2024}\natexlab{b}.
\newblock \showarticletitle{Universal Humanoid Motion Representations for Physics-Based Control}. In \bibinfo{booktitle}{\emph{Proceedings of the International Conference on Learning Representations (ICLR)}}.
\newblock


\bibitem[Luo et~al\mbox{.}(2023)]%
        {Luo2023PerpetualHC}
\bibfield{author}{\bibinfo{person}{Zhengyi Luo}, \bibinfo{person}{Jinkun Cao}, \bibinfo{person}{Alexander~W. Winkler}, \bibinfo{person}{Kris Kitani}, {and} \bibinfo{person}{Weipeng Xu}.} \bibinfo{year}{2023}\natexlab{}.
\newblock \showarticletitle{Perpetual Humanoid Control for Real-Time Simulated Avatars}. In \bibinfo{booktitle}{\emph{Proceedings of the IEEE/CVF International Conference on Computer Vision (ICCV)}}.
\newblock


\bibitem[Luo et~al\mbox{.}(2022)]%
        {Luo2022EmbodiedSH}
\bibfield{author}{\bibinfo{person}{Zhengyi Luo}, \bibinfo{person}{Shun Iwase}, \bibinfo{person}{Ye Yuan}, {and} \bibinfo{person}{Kris Kitani}.} \bibinfo{year}{2022}\natexlab{}.
\newblock \showarticletitle{Embodied Scene-aware Human Pose Estimation}. In \bibinfo{booktitle}{\emph{Advances in Neural Information Processing Systems (NeurIPS)}}.
\newblock


\bibitem[Mahmood et~al\mbox{.}(2019)]%
        {AMASS:ICCV:2019}
\bibfield{author}{\bibinfo{person}{Naureen Mahmood}, \bibinfo{person}{Nima Ghorbani}, \bibinfo{person}{Nikolaus~F. Troje}, \bibinfo{person}{Gerard Pons-Moll}, {and} \bibinfo{person}{Michael~J. Black}.} \bibinfo{year}{2019}\natexlab{}.
\newblock \showarticletitle{{AMASS}: Archive of Motion Capture as Surface Shapes}. In \bibinfo{booktitle}{\emph{Proceedings of the IEEE/CVF International Conference on Computer Vision (ICCV)}}.
\newblock


\bibitem[Makoviychuk et~al\mbox{.}(2021)]%
        {makoviychuk2021isaac}
\bibfield{author}{\bibinfo{person}{Viktor Makoviychuk}, \bibinfo{person}{Lukasz Wawrzyniak}, \bibinfo{person}{Yunrong Guo}, \bibinfo{person}{Michelle Lu}, \bibinfo{person}{Kier Storey}, \bibinfo{person}{Miles Macklin}, \bibinfo{person}{David Hoeller}, \bibinfo{person}{Nikita Rudin}, \bibinfo{person}{Arthur Allshire}, \bibinfo{person}{Ankur Handa}, {and} \bibinfo{person}{Gavriel State}.} \bibinfo{year}{2021}\natexlab{}.
\newblock \showarticletitle{Isaac Gym: High Performance {GPU} Based Physics Simulation For Robot Learning}. In \bibinfo{booktitle}{\emph{Advances in Neural Information Processing Systems (NeurIPS), Datasets and Benchmarks Track}}.
\newblock


\bibitem[Osman et~al\mbox{.}(2020)]%
        {STAR:2020}
\bibfield{author}{\bibinfo{person}{Ahmed A.~A. Osman}, \bibinfo{person}{Timo Bolkart}, {and} \bibinfo{person}{Michael~J. Black}.} \bibinfo{year}{2020}\natexlab{}.
\newblock \showarticletitle{{STAR}: A Sparse Trained Articulated Human Body Regressor}. In \bibinfo{booktitle}{\emph{Proceedings of the European Conference on Computer Vision (ECCV)}}.
\newblock


\bibitem[Pavlakos et~al\mbox{.}(2019)]%
        {SMPL-X:2019}
\bibfield{author}{\bibinfo{person}{Georgios Pavlakos}, \bibinfo{person}{Vasileios Choutas}, \bibinfo{person}{Nima Ghorbani}, \bibinfo{person}{Timo Bolkart}, \bibinfo{person}{Ahmed A.~A. Osman}, \bibinfo{person}{Dimitrios Tzionas}, {and} \bibinfo{person}{Michael~J. Black}.} \bibinfo{year}{2019}\natexlab{}.
\newblock \showarticletitle{Expressive Body Capture: 3D Hands, Face, and Body from a Single Image}. In \bibinfo{booktitle}{\emph{Proceedings of the IEEE/CVF Conference on Computer Vision and Pattern Recognition (CVPR)}}.
\newblock


\bibitem[Peng et~al\mbox{.}(2021b)]%
        {peng2021neural}
\bibfield{author}{\bibinfo{person}{Sida Peng}, \bibinfo{person}{Yuanqing Zhang}, \bibinfo{person}{Yinghao Xu}, \bibinfo{person}{Qianqian Wang}, \bibinfo{person}{Qing Shuai}, \bibinfo{person}{Hujun Bao}, {and} \bibinfo{person}{Xiaowei Zhou}.} \bibinfo{year}{2021}\natexlab{b}.
\newblock \showarticletitle{Neural Body: Implicit Neural Representations with Structured Latent Codes for Novel View Synthesis of Dynamic Humans}. In \bibinfo{booktitle}{\emph{Proceedings of the IEEE/CVF Conference on Computer Vision and Pattern Recognition (CVPR)}}.
\newblock


\bibitem[Peng et~al\mbox{.}(2018a)]%
        {2018-TOG-deepMimic}
\bibfield{author}{\bibinfo{person}{Xue~Bin Peng}, \bibinfo{person}{Pieter Abbeel}, \bibinfo{person}{Sergey Levine}, {and} \bibinfo{person}{Michiel van~de Panne}.} \bibinfo{year}{2018}\natexlab{a}.
\newblock \showarticletitle{DeepMimic: Example-guided Deep Reinforcement Learning of Physics-based Character Skills}.
\newblock \bibinfo{journal}{\emph{ACM Transactions on Graphics (TOG)}} (\bibinfo{year}{2018}).
\newblock


\bibitem[Peng et~al\mbox{.}(2022)]%
        {2022-TOG-ASE}
\bibfield{author}{\bibinfo{person}{Xue~Bin Peng}, \bibinfo{person}{Yunrong Guo}, \bibinfo{person}{Lina Halper}, \bibinfo{person}{Sergey Levine}, {and} \bibinfo{person}{Sanja Fidler}.} \bibinfo{year}{2022}\natexlab{}.
\newblock \showarticletitle{ASE: Large-Scale Reusable Adversarial Skill Embeddings for Physically Simulated Characters}.
\newblock \bibinfo{journal}{\emph{ACM Transactions on Graphics (TOG)}} (\bibinfo{year}{2022}).
\newblock


\bibitem[Peng et~al\mbox{.}(2018b)]%
        {peng2018sfv}
\bibfield{author}{\bibinfo{person}{Xue~Bin Peng}, \bibinfo{person}{Angjoo Kanazawa}, \bibinfo{person}{Jitendra Malik}, \bibinfo{person}{Pieter Abbeel}, {and} \bibinfo{person}{Sergey Levine}.} \bibinfo{year}{2018}\natexlab{b}.
\newblock \showarticletitle{SFV: Reinforcement Learning of Physical Skills from Videos}.
\newblock \bibinfo{journal}{\emph{ACM Transactions on Graphics (TOG)}} (\bibinfo{year}{2018}).
\newblock


\bibitem[Peng et~al\mbox{.}(2021a)]%
        {AMP}
\bibfield{author}{\bibinfo{person}{Xue~Bin Peng}, \bibinfo{person}{Ze Ma}, \bibinfo{person}{Pieter Abbeel}, \bibinfo{person}{Sergey Levine}, {and} \bibinfo{person}{Angjoo Kanazawa}.} \bibinfo{year}{2021}\natexlab{a}.
\newblock \showarticletitle{AMP: adversarial motion priors for stylized physics-based character control}.
\newblock \bibinfo{journal}{\emph{ACM Transactions on Graphics (TOG)}} (\bibinfo{year}{2021}).
\newblock


\bibitem[Rajasegaran et~al\mbox{.}(2022)]%
        {PHALP}
\bibfield{author}{\bibinfo{person}{Jathushan Rajasegaran}, \bibinfo{person}{Georgios Pavlakos}, \bibinfo{person}{Angjoo Kanazawa}, {and} \bibinfo{person}{Jitendra Malik}.} \bibinfo{year}{2022}\natexlab{}.
\newblock \showarticletitle{Tracking People by Predicting 3D Appearance, Location \& Pose}. In \bibinfo{booktitle}{\emph{Proceedings of the IEEE/CVF Conference on Computer Vision and Pattern Recognition (CVPR)}}.
\newblock


\bibitem[Rempe et~al\mbox{.}(2021)]%
        {rempe2021humor}
\bibfield{author}{\bibinfo{person}{Davis Rempe}, \bibinfo{person}{Tolga Birdal}, \bibinfo{person}{Aaron Hertzmann}, \bibinfo{person}{Jimei Yang}, \bibinfo{person}{Srinath Sridhar}, {and} \bibinfo{person}{Leonidas~J. Guibas}.} \bibinfo{year}{2021}\natexlab{}.
\newblock \showarticletitle{HuMoR: 3D Human Motion Model for Robust Pose Estimation}. In \bibinfo{booktitle}{\emph{Proceedings of the IEEE/CVF International Conference on Computer Vision (ICCV)}}.
\newblock


\bibitem[Shen et~al\mbox{.}(2024)]%
        {shen2024gvhmr}
\bibfield{author}{\bibinfo{person}{Zehong Shen}, \bibinfo{person}{Huaijin Pi}, \bibinfo{person}{Yan Xia}, \bibinfo{person}{Zhi Cen}, \bibinfo{person}{Sida Peng}, \bibinfo{person}{Zechen Hu}, \bibinfo{person}{Hujun Bao}, \bibinfo{person}{Ruizhen Hu}, {and} \bibinfo{person}{Xiaowei Zhou}.} \bibinfo{year}{2024}\natexlab{}.
\newblock \showarticletitle{World-Grounded Human Motion Recovery via Gravity-View Coordinates}. In \bibinfo{booktitle}{\emph{SIGGRAPH Asia Conference Papers (SA Conference Papers)}}.
\newblock


\bibitem[Shimada et~al\mbox{.}(2020)]%
        {shimada2020physcap}
\bibfield{author}{\bibinfo{person}{Soshi Shimada}, \bibinfo{person}{Vladislav Golyanik}, \bibinfo{person}{Weipeng Xu}, {and} \bibinfo{person}{Christian Theobalt}.} \bibinfo{year}{2020}\natexlab{}.
\newblock \showarticletitle{PhysCap: Physically Plausible Monocular 3D Motion Capture in Real Time}.
\newblock \bibinfo{journal}{\emph{ACM Transactions on Graphics (TOG)}} (\bibinfo{year}{2020}).
\newblock


\bibitem[Shin et~al\mbox{.}(2024)]%
        {WHAM}
\bibfield{author}{\bibinfo{person}{Soyong Shin}, \bibinfo{person}{Juyong Kim}, \bibinfo{person}{Eni Halilaj}, {and} \bibinfo{person}{Michael~J. Black}.} \bibinfo{year}{2024}\natexlab{}.
\newblock \showarticletitle{WHAM: Reconstructing World-grounded Humans with Accurate 3D Motion}. In \bibinfo{booktitle}{\emph{Proceedings of the IEEE/CVF Conference on Computer Vision and Pattern Recognition (CVPR)}}.
\newblock


\bibitem[Sun et~al\mbox{.}(2023)]%
        {TRACE}
\bibfield{author}{\bibinfo{person}{Yu Sun}, \bibinfo{person}{Qian Bao}, \bibinfo{person}{Wu Liu}, \bibinfo{person}{Tao Mei}, {and} \bibinfo{person}{Michael~J. Black}.} \bibinfo{year}{2023}\natexlab{}.
\newblock \showarticletitle{TRACE: 5D Temporal Regression of Avatars with Dynamic Cameras in 3D Environments}. In \bibinfo{booktitle}{\emph{Proceedings of the IEEE/CVF Conference on Computer Vision and Pattern Recognition (CVPR)}}.
\newblock


\bibitem[Teed et~al\mbox{.}(2023)]%
        {NEURIPS2023_7ac484b0}
\bibfield{author}{\bibinfo{person}{Zachary Teed}, \bibinfo{person}{Lahav Lipson}, {and} \bibinfo{person}{Jia Deng}.} \bibinfo{year}{2023}\natexlab{}.
\newblock \showarticletitle{Deep Patch Visual Odometry}. In \bibinfo{booktitle}{\emph{Advances in Neural Information Processing Systems (NeurIPS)}}.
\newblock


\bibitem[Tessler et~al\mbox{.}(2024)]%
        {tessler2024maskedmimic}
\bibfield{author}{\bibinfo{person}{Chen Tessler}, \bibinfo{person}{Yunrong Guo}, \bibinfo{person}{Ofir Nabati}, \bibinfo{person}{Gal Chechik}, {and} \bibinfo{person}{Xue~Bin Peng}.} \bibinfo{year}{2024}\natexlab{}.
\newblock \showarticletitle{MaskedMimic: Unified Physics-Based Character Control Through Masked Motion Inpainting}.
\newblock \bibinfo{journal}{\emph{ACM Transactions on Graphics (TOG)}} (\bibinfo{year}{2024}).
\newblock


\bibitem[Tessler et~al\mbox{.}(2023)]%
        {tessler2023calm}
\bibfield{author}{\bibinfo{person}{Chen Tessler}, \bibinfo{person}{Yoni Kasten}, \bibinfo{person}{Yunrong Guo}, \bibinfo{person}{Shie Mannor}, \bibinfo{person}{Gal Chechik}, {and} \bibinfo{person}{Xue~Bin Peng}.} \bibinfo{year}{2023}\natexlab{}.
\newblock \showarticletitle{CALM: Conditional Adversarial Latent Models for Directable Virtual Characters}.
\newblock \bibinfo{journal}{\emph{ACM Transactions on Graphics (TOG)}} (\bibinfo{year}{2023}).
\newblock


\bibitem[Todorov et~al\mbox{.}(2012)]%
        {todorov2012mujoco}
\bibfield{author}{\bibinfo{person}{Emanuel Todorov}, \bibinfo{person}{Tom Erez}, {and} \bibinfo{person}{Yuval Tassa}.} \bibinfo{year}{2012}\natexlab{}.
\newblock \showarticletitle{MuJoCo: A Physics Engine for Model-Based Control}. In \bibinfo{booktitle}{\emph{Proceedings of the IEEE/RSJ International Conference on Intelligent Robots and Systems (IROS)}}.
\newblock


\bibitem[Tripathi et~al\mbox{.}(2023)]%
        {tripathi2023intuitivephysics}
\bibfield{author}{\bibinfo{person}{Shashank Tripathi}, \bibinfo{person}{Lea M{\"u}ller}, \bibinfo{person}{Chun-Hao~P. Huang}, \bibinfo{person}{Omid Taheri}, \bibinfo{person}{Michael~J. Black}, {and} \bibinfo{person}{Dimitrios Tzionas}.} \bibinfo{year}{2023}\natexlab{}.
\newblock \showarticletitle{3D Human Pose Estimation via Intuitive Physics}. In \bibinfo{booktitle}{\emph{Proceedings of the IEEE/CVF Conference on Computer Vision and Pattern Recognition (CVPR)}}.
\newblock


\bibitem[Wagener et~al\mbox{.}(2022)]%
        {wagener2022mocapact}
\bibfield{author}{\bibinfo{person}{Nolan Wagener}, \bibinfo{person}{Andrey Kolobov}, \bibinfo{person}{Felipe~Vieira Frujeri}, \bibinfo{person}{Ricky Loynd}, \bibinfo{person}{Ching-An Cheng}, {and} \bibinfo{person}{Matthew Hausknecht}.} \bibinfo{year}{2022}\natexlab{}.
\newblock \showarticletitle{{MoCapAct}: A Multi-Task Dataset for Simulated Humanoid Control}. In \bibinfo{booktitle}{\emph{Advances in Neural Information Processing Systems (NeurIPS)}}.
\newblock


\bibitem[Wang et~al\mbox{.}(2024a)]%
        {TRAM}
\bibfield{author}{\bibinfo{person}{Yufu Wang}, \bibinfo{person}{Ziyun Wang}, \bibinfo{person}{Lingjie Liu}, {and} \bibinfo{person}{Kostas Daniilidis}.} \bibinfo{year}{2024}\natexlab{a}.
\newblock \showarticletitle{TRAM: Global Trajectory and Motion of 3D Humans from in-the-wild Videos}. In \bibinfo{booktitle}{\emph{Proceedings of the European Conference on Computer Vision (ECCV)}}.
\newblock


\bibitem[Wang et~al\mbox{.}(2024b)]%
        {wang2024skillmimic}
\bibfield{author}{\bibinfo{person}{Yinhuai Wang}, \bibinfo{person}{Qihan Zhao}, \bibinfo{person}{Runyi Yu}, \bibinfo{person}{Ailing Zeng}, \bibinfo{person}{Jing Lin}, \bibinfo{person}{Zhengyi Luo}, \bibinfo{person}{Hok~Wai Tsui}, \bibinfo{person}{Jiwen Yu}, \bibinfo{person}{Xiu Li}, \bibinfo{person}{Qifeng Chen}, \bibinfo{person}{Jian Zhang}, \bibinfo{person}{Lei Zhang}, {and} \bibinfo{person}{Ping Tan}.} \bibinfo{year}{2024}\natexlab{b}.
\newblock \bibinfo{title}{SkillMimic: Learning Reusable Basketball Skills from Demonstrations}.
\newblock
\showeprint[arxiv]{2408.15270v1}~[cs.CV]


\bibitem[Winkler et~al\mbox{.}(2022a)]%
        {winkler2022questsim}
\bibfield{author}{\bibinfo{person}{Alexander Winkler}, \bibinfo{person}{Jungdam Won}, {and} \bibinfo{person}{Yuting Ye}.} \bibinfo{year}{2022}\natexlab{a}.
\newblock \showarticletitle{QuestSim: Human Motion Tracking from Sparse Sensors with Simulated Avatars}. In \bibinfo{booktitle}{\emph{SIGGRAPH Asia Conference Papers (SA Conference Papers)}}.
\newblock


\bibitem[Winkler et~al\mbox{.}(2022b)]%
        {10.1145/3550469.3555411}
\bibfield{author}{\bibinfo{person}{Alexander Winkler}, \bibinfo{person}{Jungdam Won}, {and} \bibinfo{person}{Yuting Ye}.} \bibinfo{year}{2022}\natexlab{b}.
\newblock \showarticletitle{QuestSim: Human Motion Tracking from Sparse Sensors with Simulated Avatars}. In \bibinfo{booktitle}{\emph{SIGGRAPH Asia Conference Papers (SA Conference Papers)}}.
\newblock


\bibitem[Xiang et~al\mbox{.}(2019)]%
        {xiang2019monocular}
\bibfield{author}{\bibinfo{person}{Donglai Xiang}, \bibinfo{person}{Hanbyul Joo}, {and} \bibinfo{person}{Yaser Sheikh}.} \bibinfo{year}{2019}\natexlab{}.
\newblock \showarticletitle{Monocular Total Capture: Posing Face, Body, and Hands in the Wild}. In \bibinfo{booktitle}{\emph{Proceedings of the IEEE/CVF Conference on Computer Vision and Pattern Recognition (CVPR)}}.
\newblock


\bibitem[Xu et~al\mbox{.}(2020)]%
        {50649}
\bibfield{author}{\bibinfo{person}{Hongyi Xu}, \bibinfo{person}{Eduard~Gabriel Bazavan}, \bibinfo{person}{Andrei Zanfir}, \bibinfo{person}{Bill Freeman}, \bibinfo{person}{Rahul Sukthankar}, {and} \bibinfo{person}{Cristian Sminchisescu}.} \bibinfo{year}{2020}\natexlab{}.
\newblock \showarticletitle{GHUM \& GHUML: Generative 3D Human Shape and Articulated Pose Models}. In \bibinfo{booktitle}{\emph{Proceedings of the IEEE/CVF Conference on Computer Vision and Pattern Recognition (CVPR)}}.
\newblock


\bibitem[Xu et~al\mbox{.}(2022)]%
        {xu2022vitpose}
\bibfield{author}{\bibinfo{person}{Yufei Xu}, \bibinfo{person}{Jing Zhang}, \bibinfo{person}{Qiming Zhang}, {and} \bibinfo{person}{Dacheng Tao}.} \bibinfo{year}{2022}\natexlab{}.
\newblock \showarticletitle{Vi{TP}ose: Simple Vision Transformer Baselines for Human Pose Estimation}. In \bibinfo{booktitle}{\emph{Advances in Neural Information Processing Systems (NeurIPS)}}.
\newblock


\bibitem[Yang et~al\mbox{.}(2023)]%
        {yang2023ppr}
\bibfield{author}{\bibinfo{person}{Gengshan Yang}, \bibinfo{person}{Shuo Yang}, \bibinfo{person}{John~Z. Zhang}, \bibinfo{person}{Zachary Manchester}, {and} \bibinfo{person}{Deva Ramanan}.} \bibinfo{year}{2023}\natexlab{}.
\newblock \showarticletitle{PPR: Physically Plausible Reconstruction from Monocular Videos}. In \bibinfo{booktitle}{\emph{Proceedings of the IEEE/CVF International Conference on Computer Vision (ICCV)}}.
\newblock


\bibitem[Ye et~al\mbox{.}(2023)]%
        {ye2023slahmr}
\bibfield{author}{\bibinfo{person}{Vickie Ye}, \bibinfo{person}{Georgios Pavlakos}, \bibinfo{person}{Jitendra Malik}, {and} \bibinfo{person}{Angjoo Kanazawa}.} \bibinfo{year}{2023}\natexlab{}.
\newblock \showarticletitle{Decoupling Human and Camera Motion from Videos in the Wild}. In \bibinfo{booktitle}{\emph{Proceedings of the IEEE/CVF Conference on Computer Vision and Pattern Recognition (CVPR)}}.
\newblock


\bibitem[Yin et~al\mbox{.}(2025)]%
        {yin2025smplest}
\bibfield{author}{\bibinfo{person}{Wanqi Yin}, \bibinfo{person}{Zhongang Cai}, \bibinfo{person}{Ruisi Wang}, \bibinfo{person}{Ailing Zeng}, \bibinfo{person}{Chen Wei}, \bibinfo{person}{Qingping Sun}, \bibinfo{person}{Haiyi Mei}, \bibinfo{person}{Yanjun Wang}, \bibinfo{person}{Hui~En Pang}, \bibinfo{person}{Mingyuan Zhang}, \bibinfo{person}{Lei Zhang}, \bibinfo{person}{Chen~Change Loy}, \bibinfo{person}{Atsushi Yamashita}, \bibinfo{person}{Lei Yang}, {and} \bibinfo{person}{Ziwei Liu}.} \bibinfo{year}{2025}\natexlab{}.
\newblock \bibinfo{title}{{SMPLest-X}: Ultimate Scaling for Expressive Human Pose and Shape Estimation}.
\newblock
\showeprint[arxiv]{2501.09782}~[cs.CV]


\bibitem[Yuan et~al\mbox{.}(2022)]%
        {yuan2022glamr}
\bibfield{author}{\bibinfo{person}{Ye Yuan}, \bibinfo{person}{Umar Iqbal}, \bibinfo{person}{Pavlo Molchanov}, \bibinfo{person}{Kris Kitani}, {and} \bibinfo{person}{Jan Kautz}.} \bibinfo{year}{2022}\natexlab{}.
\newblock \showarticletitle{GLAMR: Global Occlusion-Aware Human Mesh Recovery with Dynamic Cameras}. In \bibinfo{booktitle}{\emph{Proceedings of the IEEE/CVF Conference on Computer Vision and Pattern Recognition (CVPR)}}.
\newblock


\bibitem[Yuan et~al\mbox{.}(2021)]%
        {yuan2021simpoe}
\bibfield{author}{\bibinfo{person}{Ye Yuan}, \bibinfo{person}{Shih-En Wei}, \bibinfo{person}{Tomas Simon}, \bibinfo{person}{Kris Kitani}, {and} \bibinfo{person}{Jason Saragih}.} \bibinfo{year}{2021}\natexlab{}.
\newblock \showarticletitle{SimPoE: Simulated Character Control for 3D Human Pose Estimation}. In \bibinfo{booktitle}{\emph{Proceedings of the IEEE/CVF Conference on Computer Vision and Pattern Recognition (CVPR)}}.
\newblock


\bibitem[Zhang et~al\mbox{.}(2024b)]%
        {zhang2024physpt}
\bibfield{author}{\bibinfo{person}{Yufei Zhang}, \bibinfo{person}{Jeffrey~O. Kephart}, \bibinfo{person}{Zijun Cui}, {and} \bibinfo{person}{Qiang Ji}.} \bibinfo{year}{2024}\natexlab{b}.
\newblock \showarticletitle{PhysPT: Physics-aware Pretrained Transformer for Estimating Human Dynamics from Monocular Videos}. In \bibinfo{booktitle}{\emph{Proceedings of the IEEE/CVF Conference on Computer Vision and Pattern Recognition (CVPR)}}.
\newblock


\bibitem[Zhang et~al\mbox{.}(2024a)]%
        {zhang2024incorporating}
\bibfield{author}{\bibinfo{person}{Yufei Zhang}, \bibinfo{person}{Jeffrey~O. Kephart}, {and} \bibinfo{person}{Qiang Ji}.} \bibinfo{year}{2024}\natexlab{a}.
\newblock \showarticletitle{Incorporating Physics Principles for Precise Human Motion Prediction}. In \bibinfo{booktitle}{\emph{Proceedings of the IEEE/CVF Winter Conference on Applications of Computer Vision (WACV)}}.
\newblock


\end{thebibliography}
\appendix

\section{About camera-to-world Transformation}

The key to linking the simulated world with the image domain is the $T^{c2w}_t$ transformation. There are multiple ways to compute it, and our method does not depend on a specific solution.

In the EMDB2 experiments, we combine TRAM and GVHMR. TRAM provides accurate SLAM-based camera trajectories, but its result may differ from the global frame by a rigid transformation. GVHMR, on the other hand, does not model the camera explicitly but produces outputs in gravity-aligned space, where the gravity direction is downward.

For moving camera,
we ignore translation and compute a global rotation by aligning the joint positions of GVHMR and TRAM (e.g., using Procrustes alignment). Once the rotation is known, we determine the translation by aligning the first-frame SMPL root position. We define the floor based on the foot position of the GVHMR result in the first frame, which is also used to initialize the humanoid in the simulator, similar to the PHC series. Small alignment errors are acceptable in practice. While the estimated ground height may contain small errors, it is acceptable in practice.

For static cameras, the process is simpler: we directly align the GVHMR first-frame gravity space result with the camera space.

\section{Dealing with Shape Variance}

In practice, the shape of a real human SMPL model can be approximated as a scaled version of the zero-shape SMPL. However, the humanoid used in the simulator always follows the zero-shape size. To resolve this mismatch, we apply a scale correction to the transformation:
\[
\hat{T}_{t}^{c2w} = T_{t}^{c2w}  \cdot \frac{1}{\text{scale}}
\]

\section{Reward Definition}

Following PHC+, we use a SMPL-based humanoid agent, which consists of 24 rigid bodies, 23 of which are actuated. The proprioceptive state $ s^{p}_{t} $ is defined as:
\begin{equation}
    s^{p}_{t} := (r_{t}, p_{t}, v_{t}, \omega_{t})
\end{equation}
where $ r_{t} $, $ p_{t} $, $ v_{t} $, and $ \omega_{t} $ are the simulated joint rotations, positions, velocities, and angular velocities, respectively. The reference state $\theta_{t} $ is defined as:
\begin{equation}
    \theta_{t} := (\hat{r}_{t+1} \ominus r_t, \hat{p}_{t+1} - p_t, \hat{v}_{t+1} - v_t, \hat{\omega}_{t+1} - \omega_t, \hat{r}_{t+1}, \hat{p}_{t+1})
\end{equation}
where $ \ominus $ denotes rotation difference. $ \hat{r}_{t+1} $, $ \hat{p}_{t+1} $, $ \hat{v}_{t+1} $, and $ \hat{\omega}_{t+1} $ represent the reference joint rotations, positions, velocities, and angular velocities, respectively. The imitation reward $R_{\text{pose}}$ is defined as:
\begin{equation}
    \begin{aligned}
            R_{\text{pose}} \ := w_{p}e^{-\lambda_{p}\|p_{t} - \hat{p}_{t}\|} + w_{r}e^{-\lambda_{r}\|r_{t} - \hat{r}_{t}\|} \\
    \ + w_{v} e^{-\lambda_{v}\|v_{t} - \hat{v}_{t}\|} + w_{\omega}e^{-\lambda_{\omega}\|\omega_{t} - \hat{\omega}_{t}\|}
    \end{aligned}
\end{equation}
where $ w_{\{ \cdot \}} $, $ \lambda_{\{ \cdot \}} $ denote the corresponding weights. We utilize the same weight settings as PHC+.

To obtain the style reward, $R_{\text{amp}}$, we train a discriminator $D(s^{p}_{t-10:t})$ jointly with the policy network to distinguish real motion sequences from those generated by the policy. The discriminator produces a scalar value based on the proprioception of the humanoid, encouraging the generation of realistic, human-like motion aligned with the motion prior.

\section{Test Data}

We exclude AIST++ sequences with high heels and ballet shoes, resulting in 326 videos from 4 camera views (totaling 1304 videos). From EMDB2, we remove sequences involving skateboards or stairs, which cannot be replicated in the simulator. The remaining sequences are:

\texttt{
P0\_09,  
P2\_19,  
P2\_20,  
P2\_24,  
P3\_27,  
P3\_28,  
P4\_35,  
P4\_36,  
P4\_37,  
P5\_40,  
P7\_55,  
P7\_61,  
P8\_65,  
P9\_79,  
P9\_80
}

In \texttt{P2\_24}, we removed frames 1650--1800 due to a step-up motion. The sequence was split into \texttt{P2\_24\_0} and \texttt{P2\_24\_1}. For \texttt{P4\_36}, we removed the first 300 frames, and for \texttt{P7\_61}, the first 600 frames (sitting or lying).


\end{document}